\documentclass[%
  onecolumn
   , hidempi
]{mpi2015-cscpreprint}
\makeatletter
\DeclareOldFontCommand{\rm}{\normalfont\rmfamily}{\mathrm}
\DeclareOldFontCommand{\sf}{\normalfont\sffamily}{\mathsf}
\DeclareOldFontCommand{\tt}{\normalfont\ttfamily}{\mathtt}
\DeclareOldFontCommand{\bf}{\normalfont\bfseries}{\mathbf}
\DeclareOldFontCommand{\it}{\normalfont\itshape}{\mathit}
\DeclareOldFontCommand{\sl}{\normalfont\slshape}{\@nomath\sl}
\DeclareOldFontCommand{\sc}{\normalfont\scshape}{\@nomath\sc}
\makeatother

\usepackage[american]{babel}
\usepackage{multirow}
\usepackage{graphicx}
\usepackage{amssymb}
\usepackage{amsthm}
\usepackage[mathscr]{eucal}

\numberwithin{equation}{section}
\numberwithin{figure}{section}
\numberwithin{table}{section}

\usepackage[software, hardware]{mymacros}
\usepackage{import}
\usepackage{xifthen}
\usepackage{pdfpages}
\usepackage{transparent}
\usepackage{cleveref}
\usepackage{sidecap}    
\usepackage{floatrow}
\usepackage{todonotes}
\usepackage{algorithm}
\usepackage[noend]{algpseudocode}

\makeatletter
\def\algbackskip{\hskip-\ALG@thistlm}
\makeatother

\usepackage{caption}
\usepackage{subcaption}
\usepackage{nicematrix}
\usepackage{arydshln}

\definecolor{lightgray}{gray}{0.9}
\usetikzlibrary{fadings,shapes.arrows,shadows}   

\tikzfading[name=arrowfading, top color=transparent!0, bottom color=transparent!95]
\tikzset{arrowfill/.style={top color= black!20, bottom color=green!50!black, general shadow={fill=black, shadow yshift=-0.8ex, path fading=arrowfading}}}
\tikzset{arrowstyle/.style={draw=black,arrowfill, single arrow,minimum height=#1, single arrow,
		single arrow head extend=.2cm,}}

\newcommand{\rk}{{\sffamily{RK4}}}

\renewcommand{\d}{\text{\sffamily{d}}}

\begin{document}
  
\title{Learning Dynamics from Noisy Measurements using Deep Learning with a Runge-Kutta Constraint}

\author[$\ast$]{Pawan Goyal}
\affil[$\ast$]{Max Planck Institute for Dynamics of Complex Technical Systems, 39106 Magdeburg, Germany.\authorcr
  \email{goyalp@mpi-magdeburg.mpg.de}, \orcid{0000-0003-3072-7780}
}
  
\author[$\dagger$]{Peter Benner}
\affil[$\dagger$]{Max Planck Institute for Dynamics of Complex Technical Systems, 39106 Magdeburg, Germany.\authorcr
  \email{benner@mpi-magdeburg.mpg.de}, \orcid{0000-0003-3362-4103}
}
  
\shorttitle{Learning Dynamics from Noisy Measurements using DL}
\shortauthor{P. Goyal, P. Benner}
\shortdate{}
  

  
\abstract{%
Measurement noise is an integral part while collecting data of a physical process. Thus,
noise removal is a necessary step to draw conclusions from these data, and it often becomes quite essential to construct dynamical models using these data.
We discuss a methodology to learn differential equation(s) using noisy and sparsely sampled measurements. In our methodology, the main  innovation can be seen in of integration of deep neural networks with a classical numerical integration method. 
Precisely, we aim at learning a neural network that implicitly represents the data
and an additional neural network that models the vector fields of the dependent variables. We
combine these two networks by enforcing the constraint that the data at the next time-steps
can be given by following a numerical integration scheme such as the fourth-order Runge-Kutta scheme.
The proposed framework to learn a model predicting the vector field is highly effective under noisy measurements. The approach can handle scenarios where dependent variables are not available at the same temporal grid. We demonstrate the effectiveness of the proposed method to learning models using data obtained from various differential  equations. The proposed approach provides a promising methodology to learn dynamic models, where the first-principle understanding remains opaque. 
}

\novelty{
	\begin{itemize}
		\item This work blends deep learning with a numerical integration scheme, namely the fourth-order Runge-Kutta scheme to learn dynamic models from noisy measurements.
	\item Two networks, one for implicit representation and the second one for describing dynamics, are combined by a Runge-Kutta scheme. 
	\item The proposed methodology is capable of handling noisy time-series data even when the dependent variables are not sampled on the same time or spatial grid.
	\end{itemize} 
}
\maketitle

\section{Introduction}\label{sec:introduction}
Uncovering dynamic models explaining physical phenomena and dynamic behaviors has been active research for centuries \footnote{For example, Isaac Newton developed his fundamental laws on the basis of measured data.}. When a model describing the underlying dynamics is available, it can be used for several engineering studies such as process design, optimization, predictions, and control. Conventional approaches based on physical laws, and empirical knowledge are often used to derive dynamical models. However, this is impenetrable for many complex systems,  e.g., understanding the Arctic ice pack dynamics, sea ice, power grids, neuroscience, or finance, to only name a few applications.. Data-driven methods to discover models have enormous potential to better understand transient behaviors in the latter cases. Furthermore, data acquired using imaging devices or sensors are contaminated with measurement noise. Therefore, systematic approaches that learn a dynamic model with proper treatment of noise are required. In this work, we discuss a deep learning-based approach to learn a dynamic model by attenuating noise with a Runge-Kutta scheme, thus allowing us to learn models quite accurately even when data are highly corrupted with measurement noise.

Data-driven methods to learn the governing equations of dynamic models have been studied for several decades, see, e.g., \cite{juang1994applied,ljung1999system,billings2013nonlinear}. Learning linear models from input-output data goes back to Ho and Kalman \cite{ho1966effective}. There have been several algorithmic developments for linear systems, for example, the eigensystem realization algorithm (ERA) \cite{juang1985eigensystem,longman1989recursive}, and Kalman filter-based approaches \cite{juang1993identification,phan1993linear,phan1992identification}. Dynamic mode decomposition (DMD) has also emerged as a promising approach to construct models from input-output data and has been widely applied in fluid dynamics applications, see, e.g., \cite{kalman1960new, schmid2010dynamic,tu2014dynamic}.  Furthermore, there has been a series of developments to learn nonlinear dynamic models. This includes, for example,  equations free modeling \cite{kevrekidis2003equation}, nonlinear regression \cite{voss1999amplitude}, dynamic modeling \cite{ye2015equation}, and automated inference of dynamics \cite{schmidt2011automated,daniels2015automated, daniels2015efficient}. Utilizing symbolic regression and an evolutionary algorithm \cite{bongard2007automated, schmidt2009distilling}, learning compact nonlinear models becomes possible. Moreover, leveraging sparsity (also known as sparse regression), several approaches have been proposed \cite{brunton2016sparse,mangan2016inferring,tran2017exact, schaeffer2020extracting, mangan2017model, morGoyB21a}. We also mention the work \cite{raissi2018hidden} that learns models using Gaussian process regression. All these methods have particular approaches to handle noise in the data. For example,  sparse regression methods, e.g., \cite{brunton2016sparse, mangan2016inferring, morGoyB21a}  often utilize smoothing methods before identifying models, and the work \cite{raissi2018hidden} handles measurement noise as data represented like a Gaussian process. 

Even though the aforementioned nonlinear modeling methods are appealing and powerful in providing analytic expressions for models, they are often built upon model hypotheses. For example, the success of sparse regression techniques relies on the fact that the nonlinear basis functions, describing the dynamics, lies in a candidate features library. For many complex dynamics, such as the melting Arctic ice, the utilization of these methods is not trivial. Thus, machine learning techniques, particularly deep learning-based ones, have emerged as powerful methods capable of expressing any complex function in a black-box manner given enough training data. Neural network-based approaches in the context of dynamical systems have been discussed in \cite{chen1990non,rico1993continuous,gonzalez1998identification,milano2002neural} decades ago. A particular type of neural networks, namely recurrent neural networks, intrinsically models sequences and is often used for forecasting \cite{lu2018attractor,pan2018long,pathak2017using,pathak2018hybrid,vlachas2018data}. Deep learning is also utilized to identify a coordinate transformation so that the dynamics in the transformed coordinates are almost linear or sparse in a high-dimensional feature basis, see, e.g., \cite{lusch2018deep, takeishi2017learning, yeung2019learning, champion2019data}. Furthermore, we mention that classical numerical schemes are incorporated with feed-forward neural networks to have discrete-time steppers for predictions, see \cite{gonzalez1998identification,raissi2018multistep,raissi2019physics,raissi2020hidden}.  The approaches in \cite{gonzalez1998identification, raissi2018multistep} can be interpreted as nonlinear autoregressive models~\cite{billings2013nonlinear}.  
A crucial feature of deep learning-based approaches that integrates numerical integration schemes is that vector fields are estimated using neural networks. Also, time-stepping is done using a numerical integration scheme. However, measurement data are often corrupted with noise, and these mentioned approaches do not perform any specific noise treatment.  The work in \cite{rudy2019deep} proposes a framework that explicitly incorporates the noise into a numerical time-stepping method. Though the approach has shown promising directions, its scalability remains ambiguous as the approach explicitly needs noise estimates and aims to decompose the signal explicitly into noise and ground truth. 

\begin{figure}[!tb]
	\includegraphics[width = \textwidth]{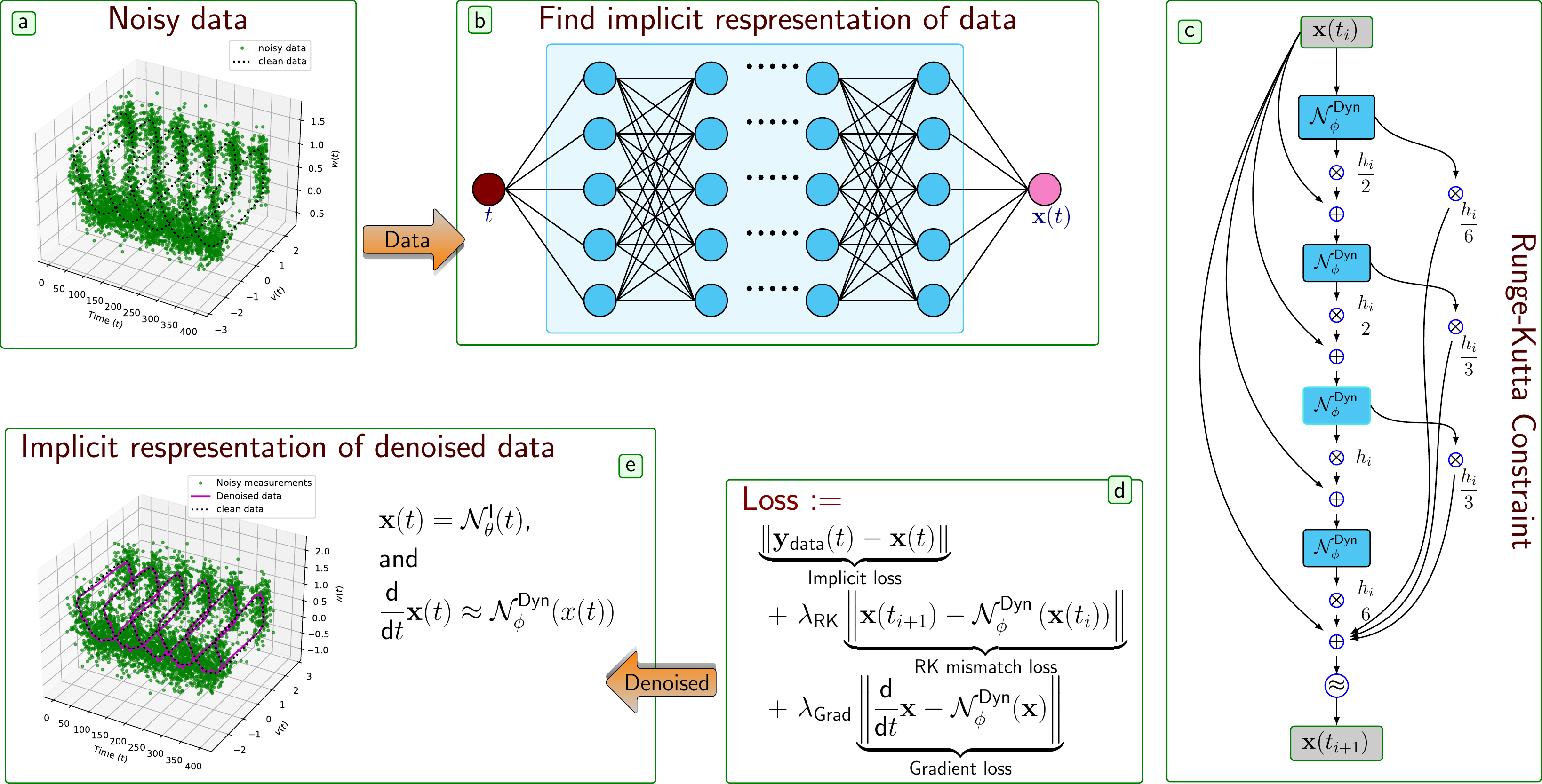}
	\caption{The figure illustrates the framework to de-noise temporal data and learns a model approximating the vector field. For this, we aim at finding an implicit representation of measurement data by a network (denoted by $\cN^{\textsf{I}}_\theta$) and another network for the vector field (denoted by $\cN^{\textsf{Dyn}}_\phi$). These two networks are connected by enforcing a Runge-Kutta scheme, shown in (c), on the output of the network $\cN^{\textsf{I}}_\theta$. Once the loss is minimized, we obtain an implicit network for de-noised data and a model $\cN^{\textsf{Dyn}}_\phi$, approximating the vector field.}
	\label{fig:method_overview}
\end{figure}

Our work introduces a framework to learn dynamics models by innovatively blending deep learning with numerical integration methods from noisy and sparse measurements. Precisely, we aim at learning two networks; one that implicitly represents given measurement data and the second one approximates the vector field; we connect these two networks by enforcing a numerical integration scheme as depicted in \Cref{fig:method_overview}. The appeal of the approach is that we do not require an explicit estimate of noise to learn a model. Furthermore, the approach is applicable even if the dependent variables are sampled on different time grids. 
The remaining structure of the paper is as follows. In \Cref{sec:method}, we present our deep learning-based framework for learning dynamics from noisy measurements by combining two networks. One of these networks implicitly represents measurement data, and the other one approximates the vector field. It is followed by connecting these two networks by enforcing a numerical integration scheme. We briefly discuss suitable architectures of neural networks for our framework in \Cref{sec:NN_arch}.  In the subsequent section, we demonstrate the effectiveness of the proposed methodology using various synthetic data with increasing levels of noise, describing various physical phenomena. We conclude the paper with a summary and future research directions. 

\section{Learning Dynamical Models using Deep Learning Constraint by  a Runge-Kutta Scheme}\label{sec:method}
Data-driven methods to learn dynamic models have flourished significantly in the last couple of decades. For these methods, quality of measurement data plays a significant role to ensure accuracy of the learned models. While dealing with real-world measurements,  sensor noise in the collected data is inevitable. Thus, before employing any data-driven method, de-noising the data is a vital step and is typically done using classical methods, e.g., smothering techniques, moving averages, or the noise is explicitly estimated along with dynamics that imposes a challenge in a large-scale setting.  In this section, we discuss our framework to learn dynamic models using noisy measurements without explicitly estimating noise. To achieve the goal, we utilize the powerful approximation capabilities of deep neural networks and its automatic differentiation feature with a numerical integration scheme. In this work, we focus on  the \emph{fourth-order Runge-Kutta} (\rk) scheme; however, the framework is flexible to use any other numerical integration scheme, or higher-order Runge-Kutta schemes. Before we proceed further, we briefly outline the \rk~scheme. For this, let us consider an autonomous nonlinear differential equation:
\begin{equation}
	\dfrac{\d}{\d t}\bx(t) = \bg(\bx(t)), \quad \bx(0) = \bx_0,
\end{equation}
where $\bx(t)\in \R^{n}$ denotes the solution at time $t$, and the continuous function $\bg(\cdot): \Rn \rightarrow \Rn$ defines the vector field. Furthermore, the solution $\bx(t_{i+1})$ can be explicitly given as follows:
\begin{equation}
	\bx(t_{i+1}) = \bx(t_{j}) + \underbrace{\int_{t_i}^{t_{i+1}}\bg(\bx(\tau)) \d\tau}_{\cG_i}. 
\end{equation}
Furthermore, we approximate the integral term $\cG_i$ using the \rk~scheme, which can be determined by a weighted sum of the vector field computed at specific locations as follows:
\begin{equation}
	\cG_i  \approx h_{i} \left(\dfrac{1}{6}\bk_1 + \dfrac{1}{3}\bk_2+ \dfrac{1}{3}\bk_3 + \dfrac{1}{6}\bk_4\right),
\end{equation}
where $h_i = t_{i+1} - t_i$, and 
\begin{align*}
	\bk_1 = \bg\left(\bx(t_i)\right) , \quad 	\bk_2 = \bg\left(\bx(t_i) + \dfrac{h_i}{2} \bk_1\right) , \quad \bk_3 = \bg\left(\bx(t_i) + \dfrac{h_i}{2} \bk_2\right) , \quad \bk_4 = \bg\left(\bx(t_i) + h_i \bk_3\right).
\end{align*}
Consequently, we can write 
\begin{equation}
	\bx(t_{i+1}) \approx \bx(t_{i})  + h_{i} \left(\dfrac{1}{6}\bk_1 + \dfrac{1}{3}\bk_2+ \dfrac{1}{3}\bk_3 + \dfrac{1}{6}\bk_4\right) =: \Pi_{\mathsf{RK}}\left(\bx(t_i)\right).
\end{equation}
In what follows,  we assume that the ground-truth (or de-noised) sequence approximately follows the \rk~steps. We emphasize that the information of the vector field at $\bx$ is directly utilized in the \rk~scheme. 

Having described the \rk~scheme, we are now ready to proceed to discuss our framework to learn dynamical models from noisy measurements by blending deep neural networks with the \rk~scheme. The approach involves two networks.  The first network implicitly represents the variable as shown in \Cref{fig:method_overview}(b), and the second network approximates the vector field, or the function $\bg(\cdot)$. These two networks are connected by attenuating the \rk~constraints. That is, the output of the implicit network is not only in the vicinity of the measurement data but also approximately follows the \rk~scheme as depicted in \Cref{fig:method_overview}(c).  To make things mathematically precise, let us denote noisy measurement data at time $t_i$ by $\by(t_i)$. Furthermore, we consider a feed-forward neural network, denoted by 
$\cN^{\textsf{I}}_{\theta}$ parameterized by $\theta$, that approximately yields an implicit representation of measurement data, i.e., 
\begin{equation}
	\cN^{\textsf{I}}_{\theta}(t_i) := \bx(t_i)    \approx \by(t_i),
\end{equation}
where $i \in \{1,\ldots, M\}$ with $M$ being the total number of measurements.  Additionally, let us denote another neural network by $\cN^{\textsf{Dyn}}_{\phi}$ parameterized by $\phi$ that approximates the vector field $\bg(\cdot)$. We connect these two networks by enforcing the output of the network $\cN^{\textsf{I}}_{\theta}$to  respects the \rk~scheme, i.e., 
\begin{equation}
	\bx(t_{i+1})  \approx \Pi_{\textsf{RK}} \bx(t_{i}), \qquad \text{and}~~
	\dfrac{\d}{\d t}  \bx(t_i) \approx \cN^{\textsf{Dyn}}_{\phi}\left(\bx(t_i)\right).
\end{equation}
As a result, our goal becomes to determine the network parameters $\{\theta, \phi\}$ such that the following loss is minimized:
\begin{equation}\label{eq:loss_func}
	\cL = \cL_{\textsf{MSE}} +  \lambda_{\textsf{RK}}\cL_{\textsf{RK}}  + \lambda_{\textsf{Grad}}  \cL_{\textsf{Grad}}, 
\end{equation}
where 
\begin{itemize}
	\item  $\cL_{\textsf{MSE}} $ denotes the root mean square error of the output of the network $\cN^{\textsf{I}}_{\theta}$ and noisy measurements, i.e., 
	\begin{equation}
		\dfrac{1}{M}\sum_{i=1}^{M} \left\|\cN^{\textsf{I}}_{\theta}(t_i) -  \by(t_i)\right\|^2.
	\end{equation}
	The loss enforces measurement data to be close to the output of the implicit network. 
	\item The term $\cL_{\mathsf{RK}}$ links the two networks by the \rk~scheme. Precisely,  the term $\cL_{\textsf{RK}}$ castigates the mismatch between  $\bx(t_{i+1})$ and $\Pi_{\textsf{RK}} \bx(t_{i})$, i.e., 
	\begin{equation}
		\dfrac{1}{M} \sum_{i=1}^{M-1}\left\|\bx(t_{i+1}) - \Pi_{\textsf{RK}}\left( \bx(t_{i})\right)\right\|^2,
	\end{equation}
	and the parameter $\lambda_{\mathsf{RK}}$ defines its weight  in the total loss.
	\item The vector field at the output of the implicit network can also be computed directly using automatic differentiation, but it also can be computed using the network~$\cN^\textsf{Dyn}_\phi$. The term $\cL_{\textsf{Grad}}$ penalizes its mismatch as follows:
	\begin{equation}
		\dfrac{1}{M} \sum_{i=1}^M\left\|\cN_\phi^{\textsf{Dyn}}(\bx (t_i))  - \dfrac{\d}{\d t} \bx(t_i) \right\|^2,
	\end{equation}
\end{itemize}
and $\lambda_{\mathsf{Grad}}$ is its corresponding regularization parameter. 

The total loss $\cL$ can be minimized using a gradient-based optimizer such as Adam \cite{kingma2014adam}. Once the networks are trained and have found their parameters that minimize the loss, we can generate the de-noised variables using the implicit network $\cN_\theta^{\textsf{I}}$, and the vector field by the network $\cN_\phi^{\textsf{Dyn}}$. Note that due to the implicit nature of the network, the measurement data can be at variable time steps, and we can estimate the solution at any arbitrary time. Moreover, we also obtain the network $\cN_\phi^{\textsf{Dyn}}(\cdot)$ that approximately provides the vector field for $\bx$; hence, one can use it to make predictions.


\section{Possible Extensions of the Approach}
In many instances, dynamical processes may involve system parameters, and by varying them, the processes exhibit different dynamics.  Also, on several occasions, dynamics are governed by underlying partial differential equations. In this section, we shortly discuss extensions of the proposed approach to these two cases. 
\subsection{Parametric models}
The approach discussed in the previous section readily extends to parametric cases. Let us consider a parametric differential equation as follows:
\begin{equation}
	\dfrac{\d}{\d t} \bx(t;\mu) = \bg(\bx(t;\mu),\mu),
\end{equation} 
where $\mu \in \Rd$ is the system parameter. To handle parameter $\mu$, we can simply take the parameter $\mu$ as an additional input to the implicit network $\cN^{\mathsf{I}}_{\theta}$ that yields the dependent variables at a given time and parameter. Furthermore, to learn the function $\bg(\bx(t;\mu), \mu)$, we take the parameter $\mu$ as an input as well along with $\bx(t;\mu)$ to obtain a parameterized dynamical model to predict the vector field at a given $\bx$ and parameter.

\subsection{Partial differential equations}
Many cases, for example, dynamics of flows, dynamical behaviors, are governed by partial differential equations; thus, the dependent variable $\bx$ is highly influenced by its neighbors. In such a case, we construct an implicit representation for measurement data such that the implicit network takes time $t$ and the spatial coordinates $\zeta$ as inputs and yields dependent variable $\bx(t,\zeta)$. Then, we compute $\bX_\zeta$ containing $\bx$ at user-specified spatial locations. This can be used to learn a dynamic model that describes dynamics at these spatial locations. Consequently, with these discussed alterations, one can employ the approach discussed in the previous section. The strength of the approach is that the collected measurement data can be at any arbitrary spatial location. These locations can also vary with time since we construct an implicit network that is independent of any structure in the collected measurements. 

\section{Suitable Neural Networks Architectures}\label{sec:NN_arch}
\begin{figure}[!tb]
	\centering
\includegraphics[width = 0.9\textwidth]{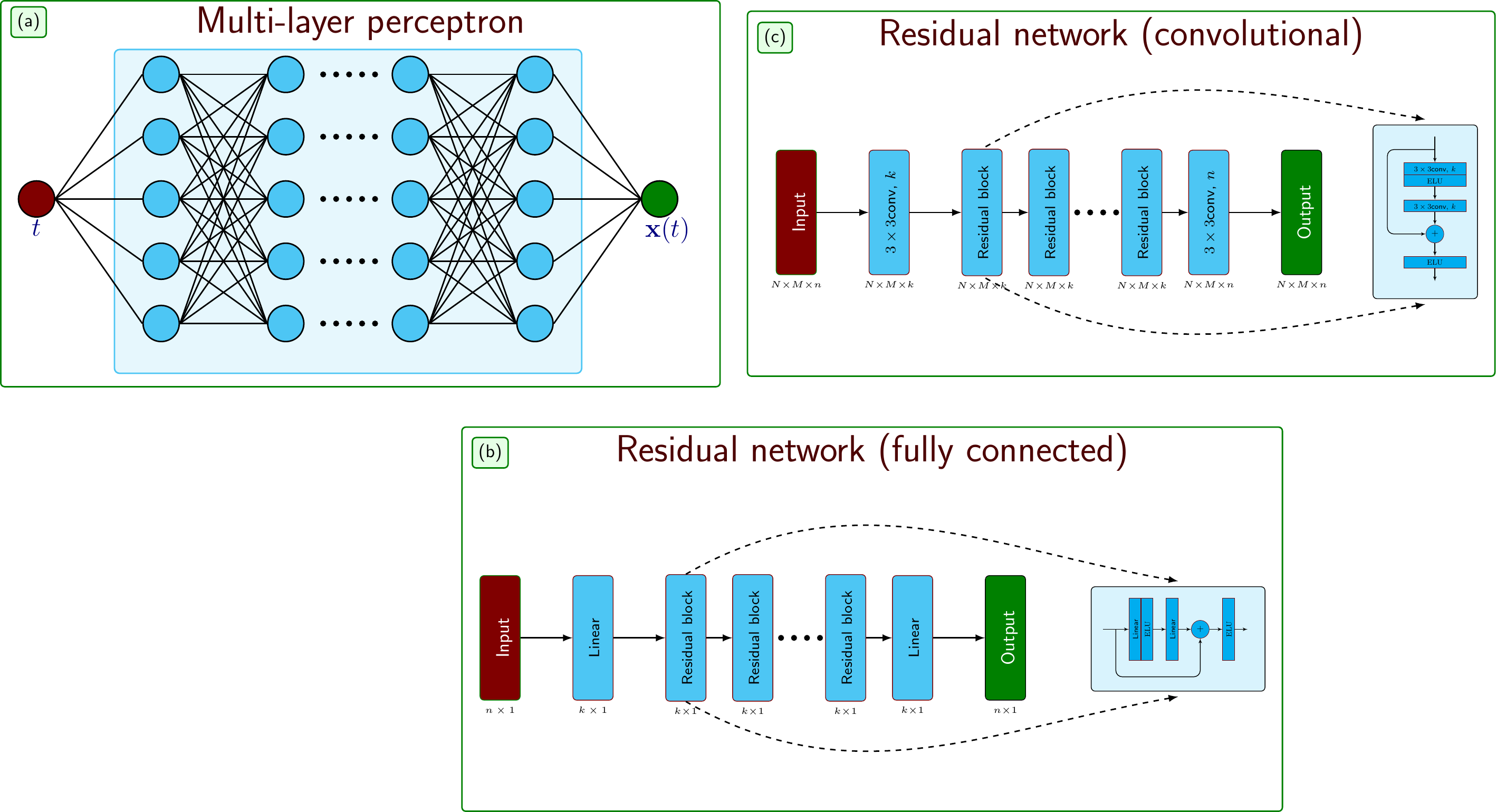}
\caption{The figure shows three potential simple architectures that can be used to learn either implicit representation or to approximate the underlying vector field. Diagram (a) is a simple multi-layer perceptron, and (b) is a residual-type network but fully connected, and (c) is a residual-type network with convolutional layers. The notation $\{3\times 3~\textsf{conv}, k\}$ means $k$ filters of $3\times 3$ receptive field. }
		\label{fig:NN_archi}
\end{figure}

Here, we briefly discuss neural network architectures suitable for our proposed approach. We require two neural networks for our framework, one for learning the implicit representation $\cN_\theta^{\textsf{I}}$ and the second one  $\cN_\theta^{\textsf{Dyn}}$ is to learn the vector field. For implicit representation, we use a fully connected multi-layer perceptron (MLP) as depicted in \Cref{fig:NN_archi}(a) with  periodic activation functions (e.g., $\sin$) \cite{sitzmann2020implicit} which has shown its ability to capture finely detailed features as well as the gradients of a function. To approximate the vector field, we consider two possibilities depending on applications. If the data do not have any spatial dependency, then we consider a simple residual-type network as illustrated in \Cref{fig:NN_archi}(b) with \emph{exponential linear unit} (ELU) as an activation function \cite{clevert2015fast}. We choose ELU as the activation function since it is continuous and differentiable and resembles a widely used activation function, namely rectified linear unit (ReLU).
On the other hand, when the data has spatial correlations, e.g., dynamics in data are governed by a partial differential equation, then it is more intuitive to use a convolutional neural network (CNN) with residual connections as depicted in \Cref{fig:NN_archi}(c). It explicitly makes use of the spatial correlation. For CNN, we also employ the batch normalization scheme \cite{ioffe2015batch} after each convolution step for a better distribution of the input to the next layer and use ELU  as an activation function.

\section{Numerical Experiments}
In this section, we investigate the performance of the approach discussed in \Cref{sec:method} to de-noise measurement data as well as learning a model for estimating the vector field. To that aim, we consider data obtained by solving several (partial) differential equations that are then corrupted using  white Gaussian noise by varying the noise level. For a given percentage of noise, we determine the noise as follows:
\begin{equation}
	\nu \sim \cN\left(0,\Sigma\right), \quad \text{with}~~ \Sigma = \mathsf{Noise}\%.
\end{equation}
We have implemented our framework using the deep learning library PyTorch \cite{paszke2019pytorch} and have optimized all networks together using the Adam optimizer \cite{kingma2014adam}. Furthermore, to train implicit networks, we map the input data to  $[-1,1]$ as recommended in \cite{sitzmann2020implicit}. Additionally, to avoid over-fitting, we add  $L_2$-regularization (also referred to as \emph{weight decay}) of the parameters of the networks and set the regularization parameter to $10^{-4}$ for all examples.  All the networks are trained for $15~000$ epochs with batch size $1$, and learning rates used to train networks are stated in each example in their respective subsections.  We have run all our experiments on a \nvidia A100 GPU.

\begin{table}[tb]
	\begin{tabular}{|c|c|c|c|c|}\hline 
		Example                           & Networks                       & Neurons & \begin{tabular}[c]{@{}c@{}}Layers or\\ residual blocks\end{tabular} & Learning rates \\ \hline 
		\multirow{2}{*}{FHN}   & For implicit representation   & 20      & 4                                                                   &  $5\cdot 10^{-4}$    \\
		& For approximating vector field & 20      & 4                                                                   & $ 10^{-3}$     \\ \hline 
		\multirow{2}{*}{Cubic oscillator} & For implicit representation   & 20      & 4                                                                   &         $5\cdot 10^{-4}$\\
		& For approximating vector field & 20      & 4                                                                   &     $ 10^{-3}$   \\ \hline 
	\end{tabular}
	\caption{The table shows the information about network architectures and learning rates.}
	\label{tab:NN_info_ODE}
\end{table}

\subsection{Fitz-Hugh Nagumo model}
In the first example, we  discuss the Fitz-Hugh Nagumo (FHN) model that explains neural dynamics in a simplistic way \cite{fitzhugh1955mathematical}. This has been used as a test case to discover the model using dictionary-based sparse regression \cite{morGoyB21a}.  The dynamics are given as follows:
\begin{equation}
	\begin{aligned}
		\bv(t) &= \bv(t) - \bw(t) - \tfrac{1}{3}\bv(t)^3 + 0.5,\\
		\bw(t) &= 0.040\bv(t) - 0.028\bw(t) + 0.032,
	\end{aligned}
\end{equation}
where $\bv(t)$ and $\bw(t)$ describe the dynamics of activation and de-activation of neurons. We collect $4~000$ measurements in time $t\in [0,400]$ at a regular interval by simulating using the initial condition $[\bv(0), \bw(0)] = [2,0]$.  We then corrupt the data artificially by adding various levels of noise.
We build two networks with the information provided in \Cref{tab:NN_info_ODE}. We have set both the parameters $\lambda_{\textsf{Dyn}}$ and $\lambda_{\textsf{Grad}}$ in the loss function \eqref{eq:loss_func} to $1$. 

\begin{figure}[!htb]
		\centering
		\includegraphics[width = 0.24\textwidth, trim = 0cm 0cm 0cm 0cm, clip ] {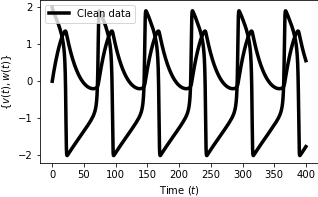}
		\includegraphics[width = 0.24\textwidth, trim = 0cm 0cm 0cm 0cm, clip ] {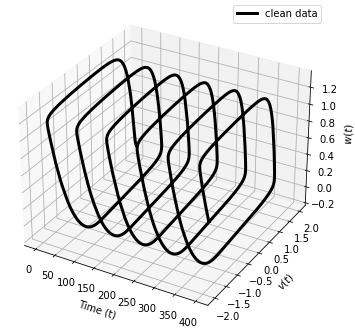}
		\includegraphics[width = 0.24\textwidth, trim = 0cm 0cm 22cm 0cm, clip ] {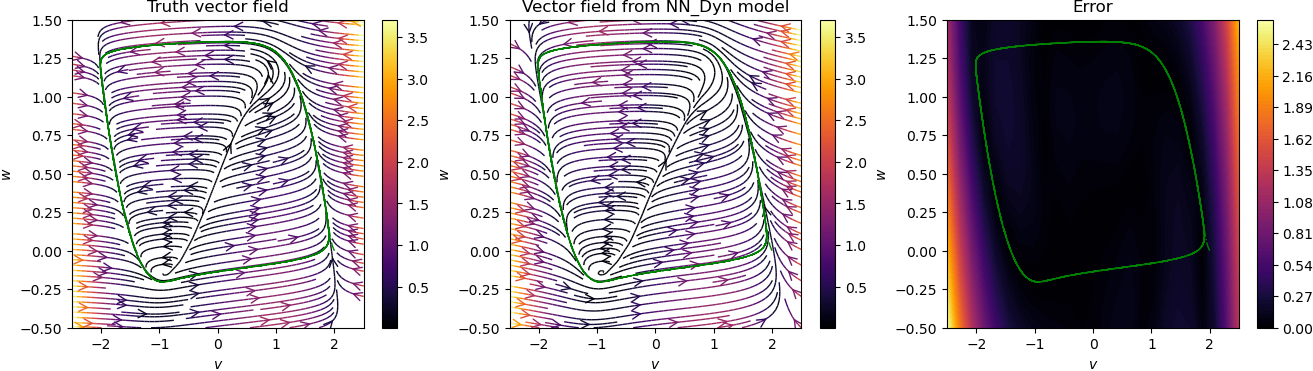}\hspace{3.8cm}
		\rotatebox{90}{\bf ~~~~~~Clean data}
		
		\includegraphics[width = 0.24\textwidth, trim = 0cm 0cm 0cm 0cm, clip ] {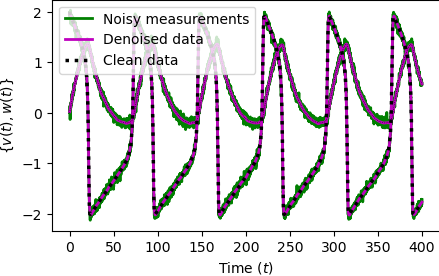}
		\includegraphics[width = 0.24\textwidth, trim = 0cm 0cm 0cm 0cm, clip ] {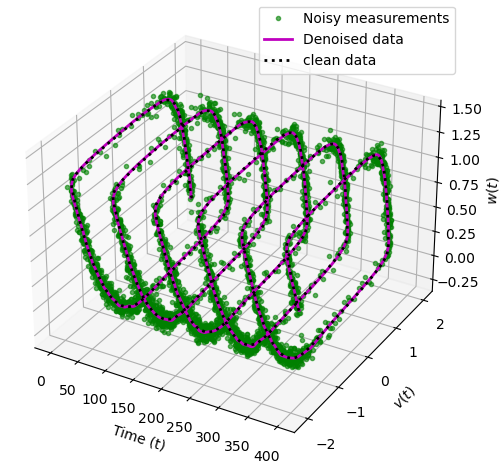}
		\includegraphics[width = 0.48\textwidth, trim = 11cm 0cm 0cm 0cm, clip ] {./Figures/FHN/FHN_vectorfield_noiselevel_0.05.png}
		\rotatebox{90}{\bf ~~~~~~Noise $5\%$}

		\includegraphics[width = 0.24\textwidth, trim = 0cm 0cm 0cm 0cm, clip ] {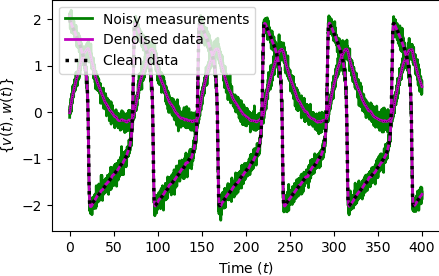}
		\includegraphics[width = 0.24\textwidth, trim = 0cm 0cm 0cm 0cm, clip ] {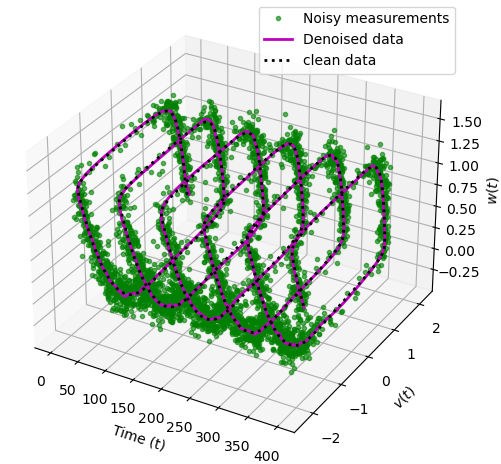}
		\includegraphics[width = 0.48\textwidth, trim = 11cm 0cm 0cm 0cm, clip ] {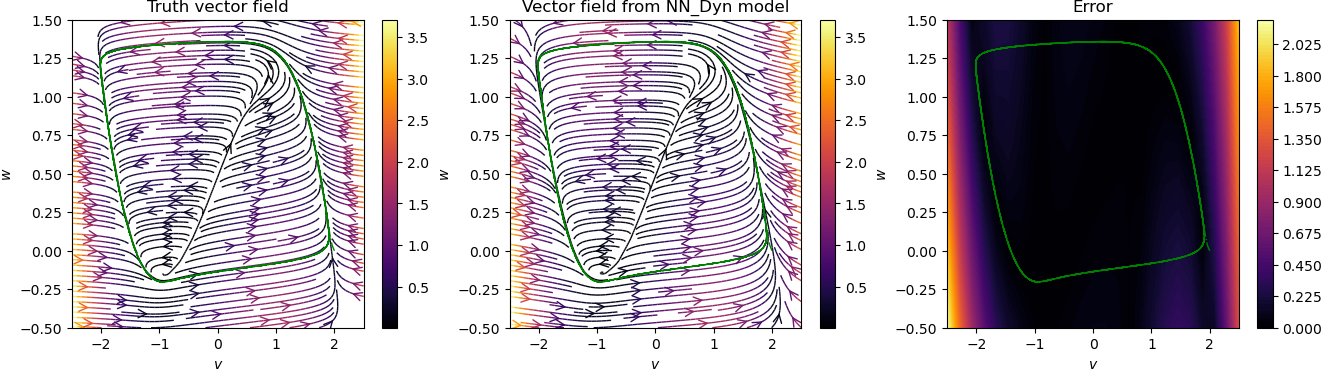}
		\rotatebox{90}{\bf ~~~~~~Noise $10\%$}

		\includegraphics[width = 0.24\textwidth, trim = 0cm 0cm 0cm 0cm, clip ] {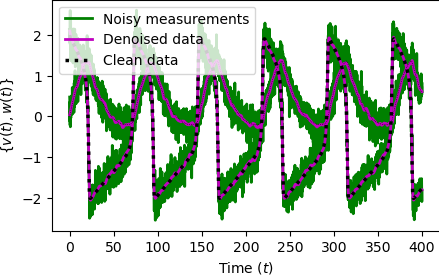}
		\includegraphics[width = 0.24\textwidth, trim = 0cm 0cm 0cm 0cm, clip ] {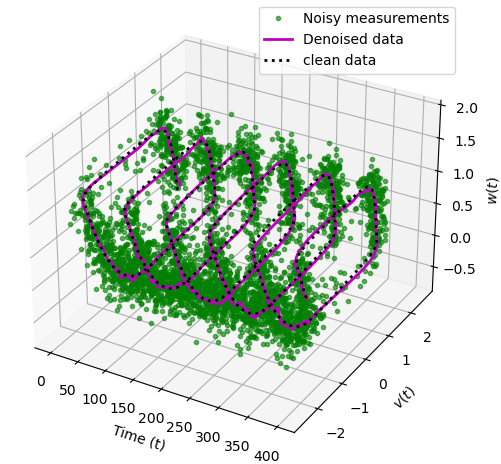}
		\includegraphics[width = 0.48\textwidth, trim = 11cm 0cm 0cm 0cm, clip ] {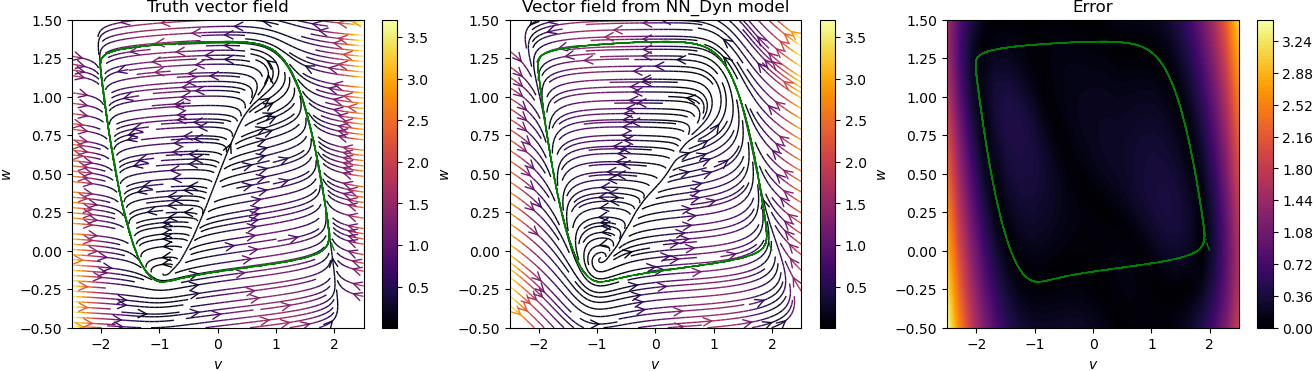}
		\rotatebox{90}{\bf ~~~~~~Noise $20\%$}
		
		\includegraphics[width = 0.24\textwidth, trim = 0cm 0cm 0cm 0cm, clip ] {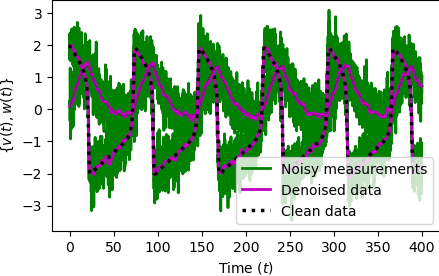}
		\includegraphics[width = 0.24\textwidth, trim = 0cm 0cm 0cm 0cm, clip ] {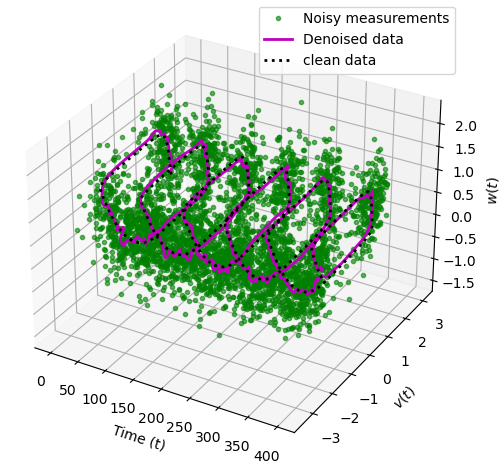}
		\includegraphics[width = 0.48\textwidth, trim = 11cm 0cm 0cm 0cm, clip ] {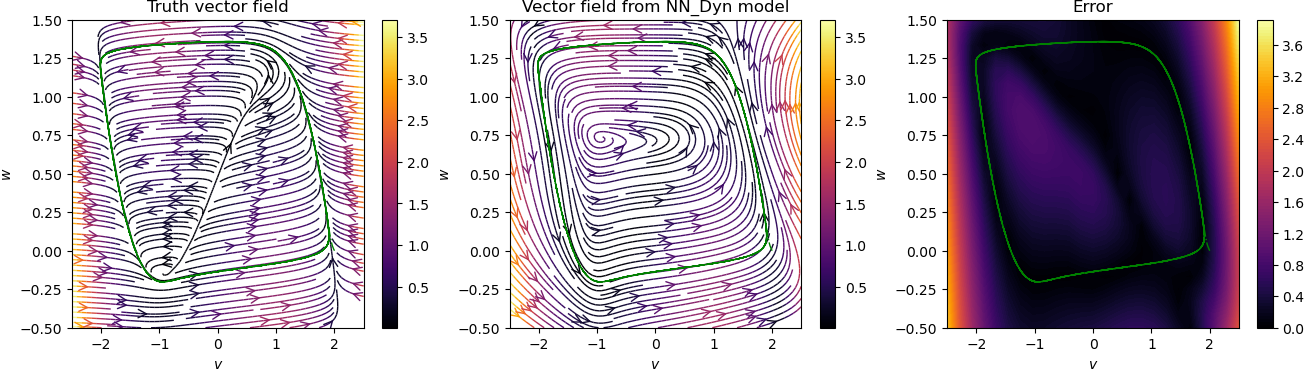}
		\rotatebox{90}{\bf ~~~~~~Noise $40\%$}

		\caption{Fitz-Hugh Nagumo model. The figure shows the performance of the proposed approach to recover the truth signal. In the first and second columns, we present  the noisy, clean, and recovered data. In the third and fourth columns, we show the estimate of the vector field using the learned neural model for it, and the green dots are the clean data in the domain. We observe that the vector fields are very well estimated in the regime of the collected data.}
		\label{fig:FHN}
\end{figure}

Having trained networks, we obtain de-noised measurement data using the implicit network and estimate the vector field using the neural network. The results are shown in \Cref{fig:FHN}. The figure demonstrates the robustness of the approach with respect to noise. The method can recover the data very close to clean data  (see the first two columns of the figure) even when the measurements are corrupted with relatively more significant noise, e.g., $20\%$ to $40\%$. Furthermore, the vector field is estimated quite accurately, at least in the regime of the collected measurements, see the third and fourth columns of the figure. However, as expected, the vector field estimates are inadequate away from the measurements, thus, showing the limitation in extrapolating to the regime where no data is available. Nevertheless, this can be improved by collecting more measurements in a different regime by varying initial conditions.

\subsection{Cubic damped model}
In the second example, we consider a damped cubic system, which is described by
\begin{equation}
	\begin{aligned}
		\dot\bx(t) &= -0.1\bx(t)^3 + 2.0\by(t)^3,\\
		\dot\by(t) &= -2.0\bx(t)^3 - 0.1\by(t)^3.
	\end{aligned}
\end{equation}
It has been one of the benchmark examples in discovering models using data, see, e.g., \cite{brunton2016discovering,morGoyB21a} but there, it is assumed that the dynamics can be given sparsely in a high-dimensional feature dictionary. Here, we do not make any such assumptions and instead learn the vector field using a neural network along the line of~\cite{rudy2019data}. For this example, we take $2~500$ data points in the time interval $[0,10]$ by simulating the model using the initial condition $[2,0]$ as done in \cite{rudy2019data}. We add various levels of noise in the clean data to have noisy measurements synthetically.    We, again, perform similar experiments as done in the previous example. We construct neural networks for implicit representation and the vector field with the parameters given in \Cref{tab:NN_info_ODE}.

\begin{figure}[!htb]
		\centering
		\includegraphics[width = 0.24\textwidth, trim = 0cm 0cm 0cm 0cm, clip ] {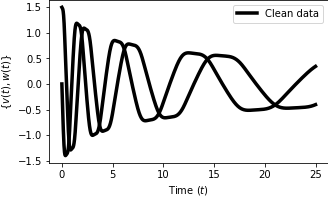}
		\includegraphics[width = 0.24\textwidth, trim = 0cm 0cm 0cm 0cm, clip ] {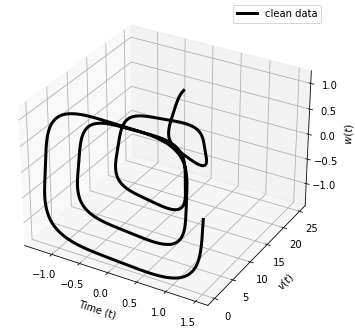}
		\includegraphics[width = 0.24\textwidth, trim = 0cm 0cm 22cm 0cm, clip ] {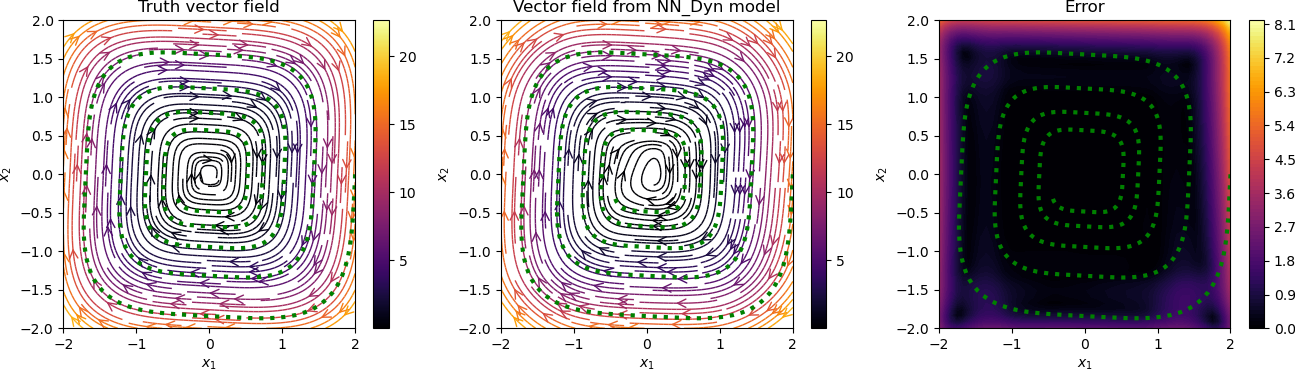}\hspace{3.8cm}
		\rotatebox{90}{\bf ~~~~~~Clean data}

		\includegraphics[width = 0.24\textwidth, trim = 0cm 0cm 0cm 0cm, clip ] {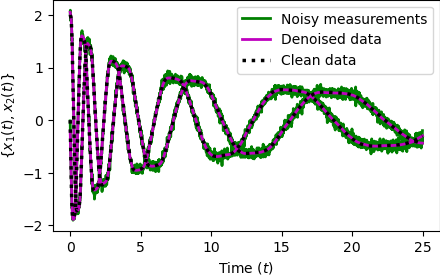}
		\includegraphics[width = 0.24\textwidth, trim = 0cm 0cm 0cm 0cm, clip ] {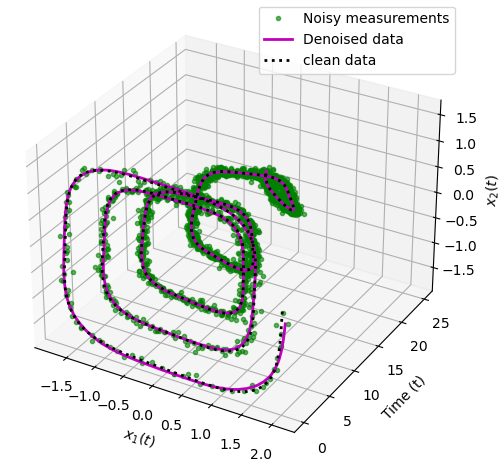}
		\includegraphics[width = 0.48\textwidth, trim = 11cm 0cm 0cm 0cm, clip ] {./Figures/Cubic/Cubic2d_vectorfield_noiselevel_0.05.png}
		\rotatebox{90}{\bf ~~~~~~Noise $5\%$}
		
		\includegraphics[width = 0.24\textwidth, trim = 0cm 0cm 0cm 0cm, clip ] {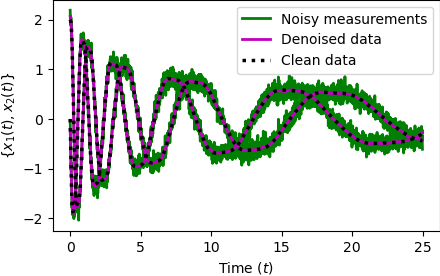}
		\includegraphics[width = 0.24\textwidth, trim = 0cm 0cm 0cm 0cm, clip ] {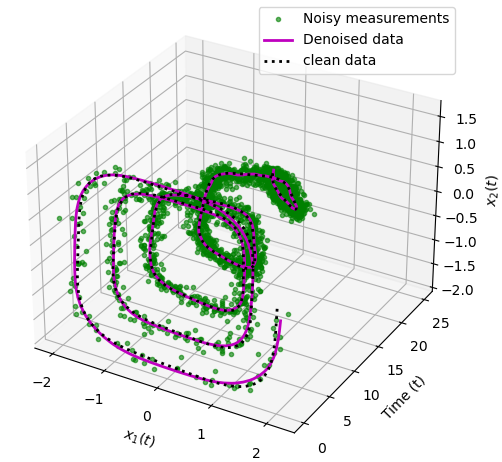}
		\includegraphics[width = 0.48\textwidth, trim = 11cm 0cm 0cm 0cm, clip ] {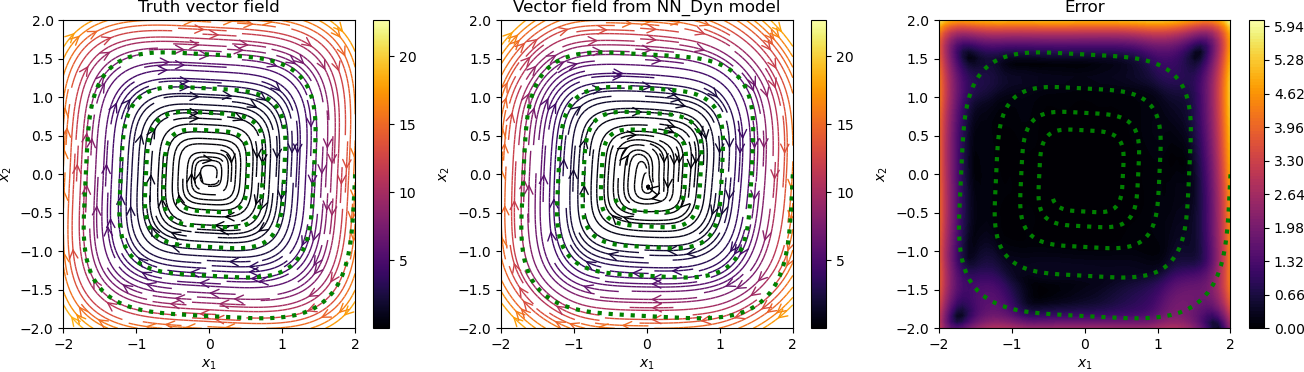}
		\rotatebox{90}{\bf ~~~~~~Noise $10\%$}
		
		\includegraphics[width = 0.24\textwidth, trim = 0cm 0cm 0cm 0cm, clip ] {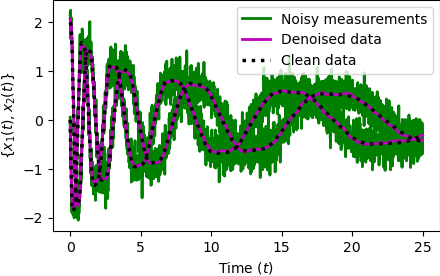}
		\includegraphics[width = 0.24\textwidth, trim = 0cm 0cm 0cm 0cm, clip ] {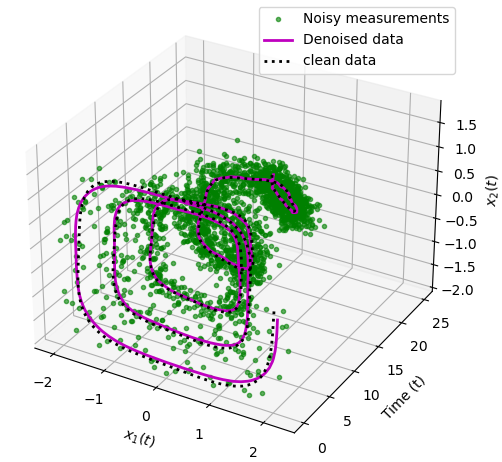}
		\includegraphics[width = 0.48\textwidth, trim = 11cm 0cm 0cm 0cm, clip ] {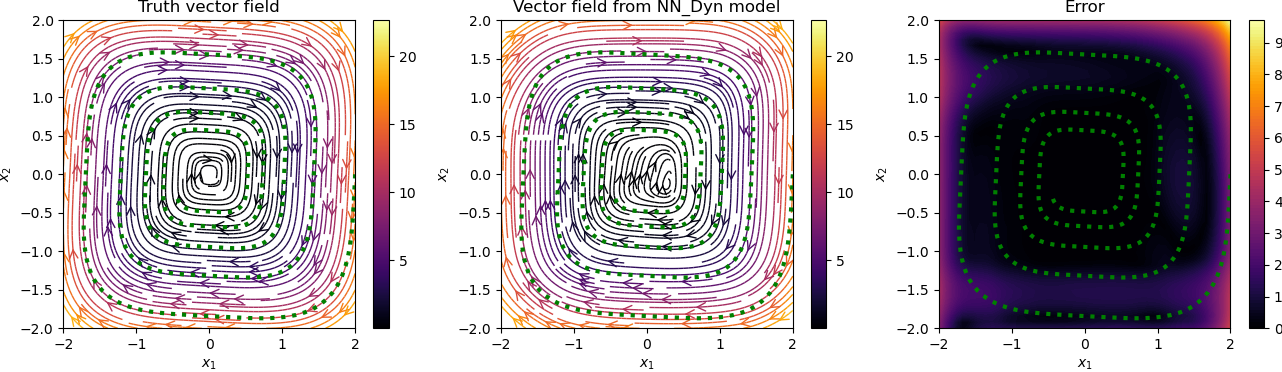}
		\rotatebox{90}{\bf ~~~~~~Noise $20\%$}
		
		\includegraphics[width = 0.24\textwidth, trim = 0cm 0cm 0cm 0cm, clip ] {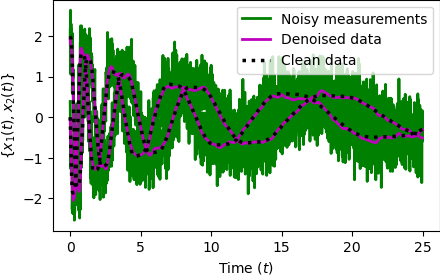}
		\includegraphics[width = 0.24\textwidth, trim = 0cm 0cm 0cm 0cm, clip ] {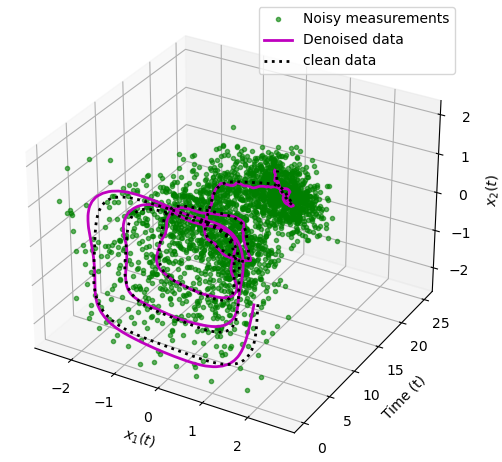}
		\includegraphics[width = 0.48\textwidth, trim = 11cm 0cm 0cm 0cm, clip ] {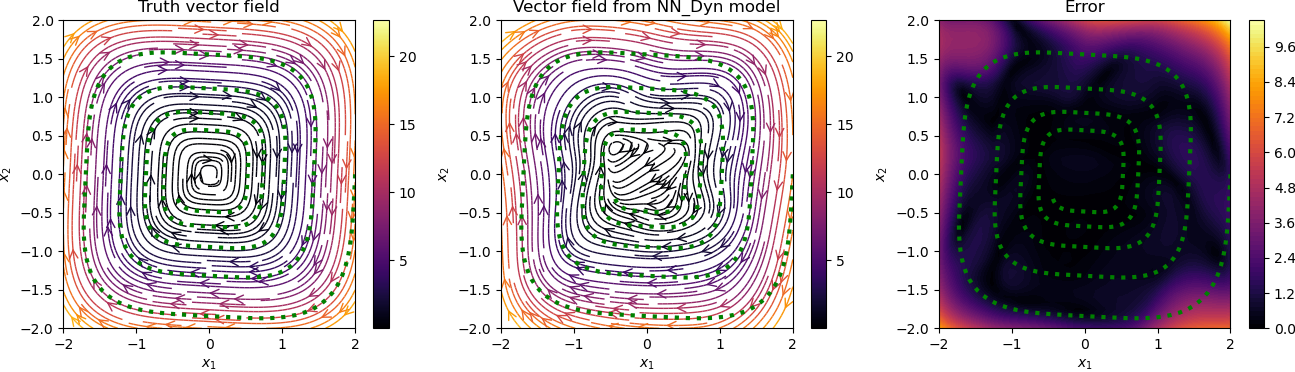}
		\rotatebox{90}{\bf ~~~~~~Noise $40\%$}
				
		\caption{Damped cubic model. We visualize the noisy, clean, and de-noised signal for various levels of noise in the measurement data (in the first two columns). We also show a comparison of the vector field obtained using the neural network with the ground truth (in the last two columns). We observe that the approach accurately  recovers the clean data from the highly noisy measurements, and accurately predicts the vector field for the model.}
		\label{fig:cubic2d}
\end{figure}

Having trained networks with parameters $\lambda_{\textsf{Dyn}} = 1$ and $\lambda_{\textsf{Grad}} = 0.05$ in the loss function \eqref{eq:loss_func}, we have an implicit network to obtain de-noised signal and a neural network approximating the vector field. We plot the results in \Cref{fig:cubic2d}, where we show noisy, clean, and de-noised data in the first two columns, and in the third and fourth columns, we plot the streamlines of the vector field, obtained using the trained neural network. We observe that the de-noised data faithfully matches with the clean data even for a high noise level, and the vector field is also close to the ground truth, at least in the region where measurement data are sampled. However, in the region where no data are available, the vector field approximation is poor, as one can expect.
However, having richer data covering a larger training regime can improve the performance of the neural network, approximating the vector field. 

\subsection{Burgers equation}
Next, we examine the case, where collected measurements have spatial correlation as well, meaning there is an underlying partial differential equation, describing the dynamics. Here, we consider a 1D viscous Burger equation. It explains several phenomenons occurring in fluid dynamics and is governed by 
\begin{equation}
	\dfrac{\partial \bu(t,\zeta)}{\partial t} = -\mu \bu_{\zeta\zeta}f(t,\zeta) + \bu(t,\zeta)\bu_{\zeta}(t,\zeta),
\end{equation}
where $\mu$ is the viscosity; $\bu_{\zeta}$ and $\bu_{\zeta\zeta}$ denote the first and second derivatives with respect to the spatial variable $\zeta$, and the equation is also subject to a boundary condition. We have taken the data from \cite{rudy2017data}, followed by artificially corrupting them  using various levels of Gaussian white noise. 
In brief, the measurements are collected at $256$ grid point in the domain $\zeta\in [-8,8]$ and at the time interval $\d t=0.1$. For more details, we refer to  \cite{rudy2017data}. 

\begin{table}[!tb]
	\begin{tabular}{|c|c|c|c|c|}\hline 
		Example                           & Networks                       & \begin{tabular}[c]{@{}c@{}} Neurons \\ or filters \end{tabular} & \begin{tabular}[c]{@{}c@{}}Layers or\\ residual blocks\end{tabular} & Learning rates \\ \hline 
		\multirow{2}{*}{Burgers example}   & For implicit representation   & 10      & 4                                                                   &  $5\cdot 10^{-4}$    \\
		& For approximating vector field & 8    & 4                                                                   & $ 10^{-3}$     \\ \hline 
		\multirow{2}{*}{  \begin{tabular}[c]{@{}c@{}}Kuramoto--Sivashinsky \\ example \end{tabular}  } & For implicit representation   & 50      & 4                                                                   &         $5\cdot 10^{-4}$\\
		& For approximating vector field & 16      & 4                                                                   &     $ 10^{-3}$   \\ \hline 
	\end{tabular}
	\caption{The table shows the information about the network for Burgers and Kuramoto--Sivashinsky examples.}
	\label{tab:NN_info_PDE}
\end{table}

\begin{figure}[!tb]
	
	\includegraphics[width = 0.25\textwidth, trim = 0cm 0cm 20cm 0cm, clip]{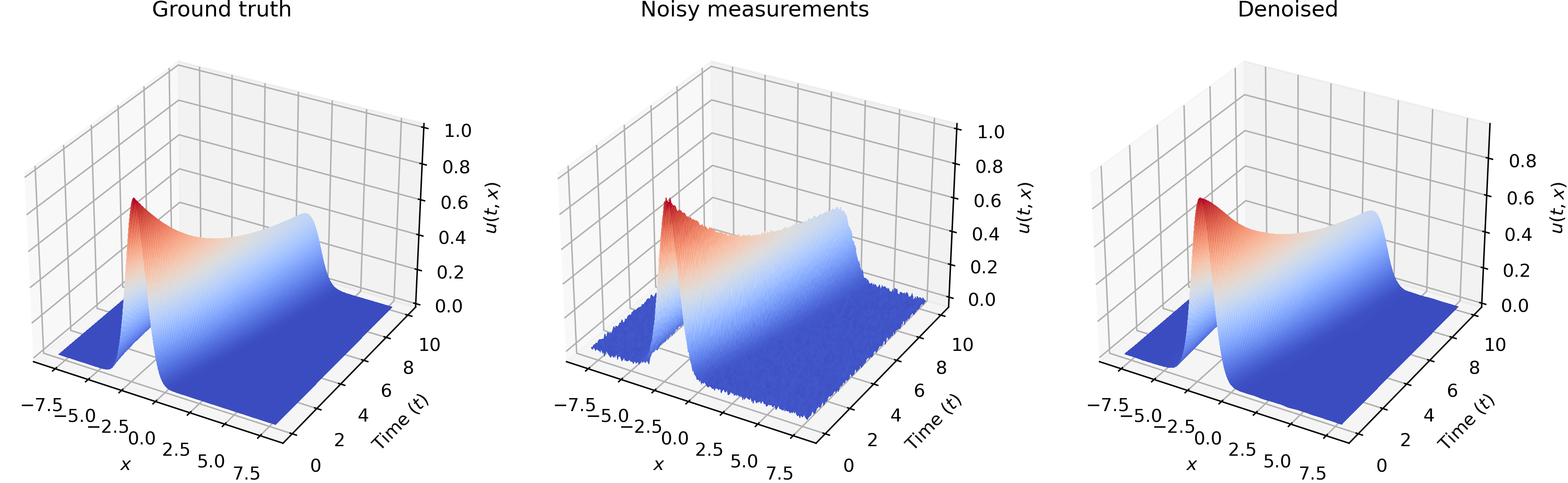}\hspace{4.4cm}
	\includegraphics[width = 0.2\textwidth, trim = 0cm 0cm 10cm 0cm, clip]{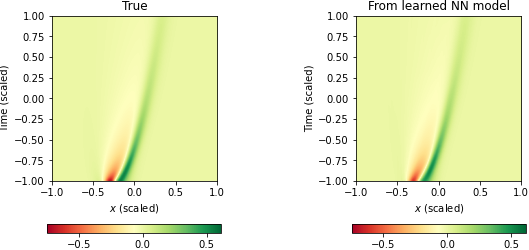} \rotatebox{90}{\hspace{1cm}clean data}
	
	\includegraphics[width = 0.5\textwidth, trim = 10cm 0cm 0cm 0cm, clip]{./Figures/Burgers/measurements_sampling1_noiselevel0.01_denoise.png}
	\includegraphics[width = 0.2\textwidth, trim = 10cm 0cm 0cm 0cm, clip]{./Figures/Burgers/measurements_sampling1_noiselevel0.01_grad}
	\rotatebox{90}{\hspace{1cm}$1\%$ noise}
	
	\includegraphics[width = 0.5\textwidth, trim = 10cm 0cm 0cm 0cm, clip]{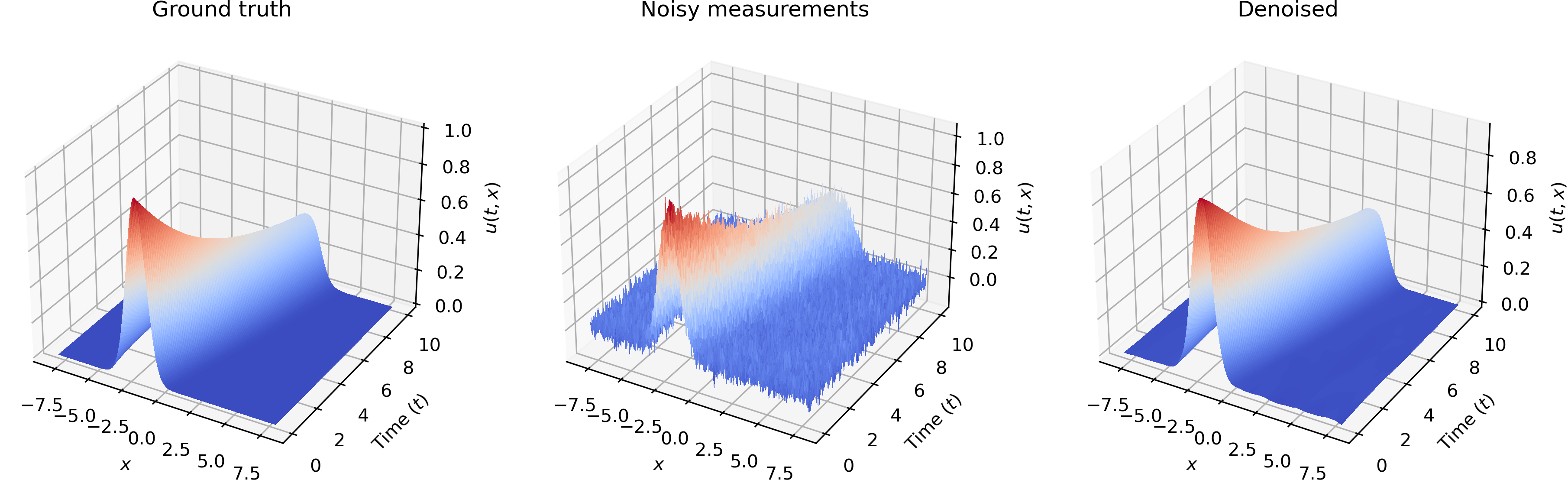}
	\includegraphics[width = 0.2\textwidth, trim = 10cm 0cm 0cm 0cm, clip]{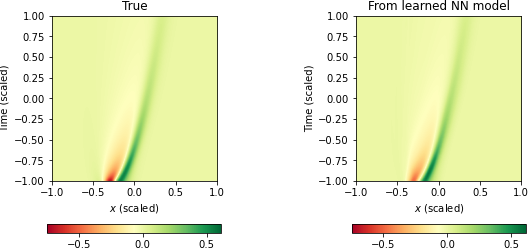}
	\rotatebox{90}{\hspace{1cm}$5\%$ noise}
	
	\includegraphics[width = 0.5\textwidth, trim = 10cm 0cm 0cm 0cm, clip]{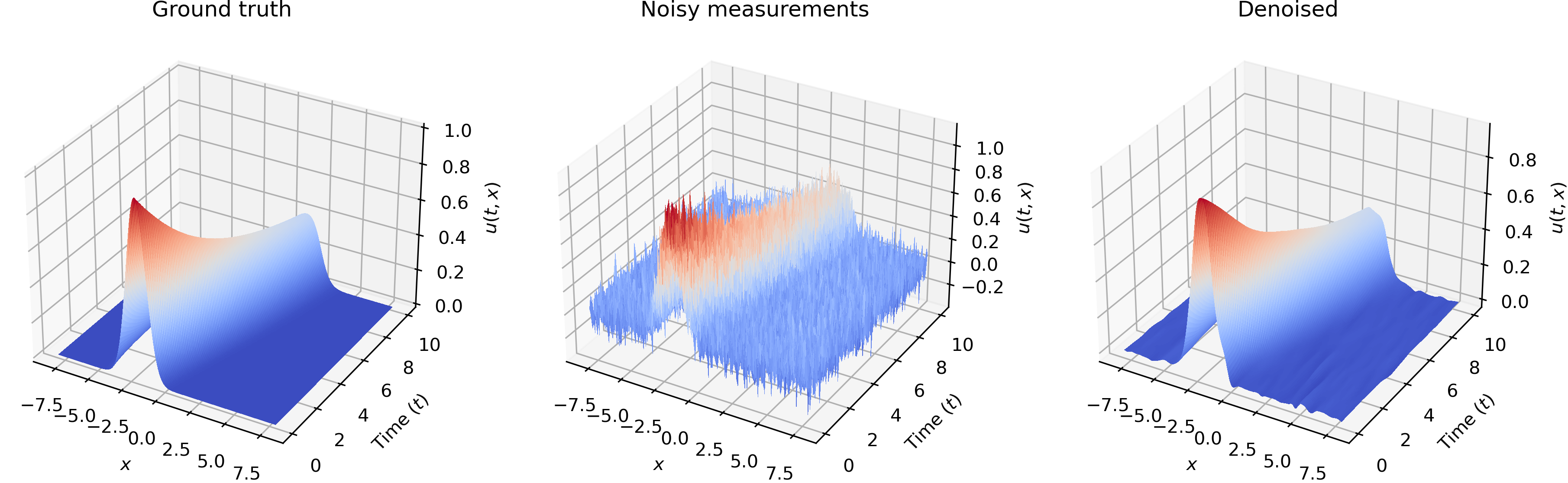}
	\includegraphics[width = 0.2\textwidth, trim = 10cm 0cm 0cm 0cm, clip]{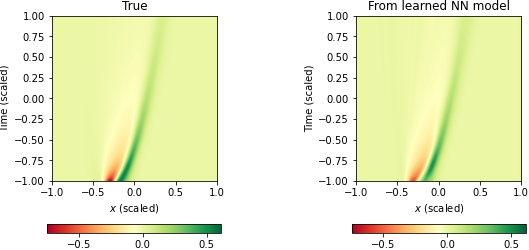}
	\rotatebox{90}{\hspace{1cm}$10\%$ noise}
	
	\includegraphics[width = 0.5\textwidth, trim = 10cm 0cm 0cm 0cm, clip]{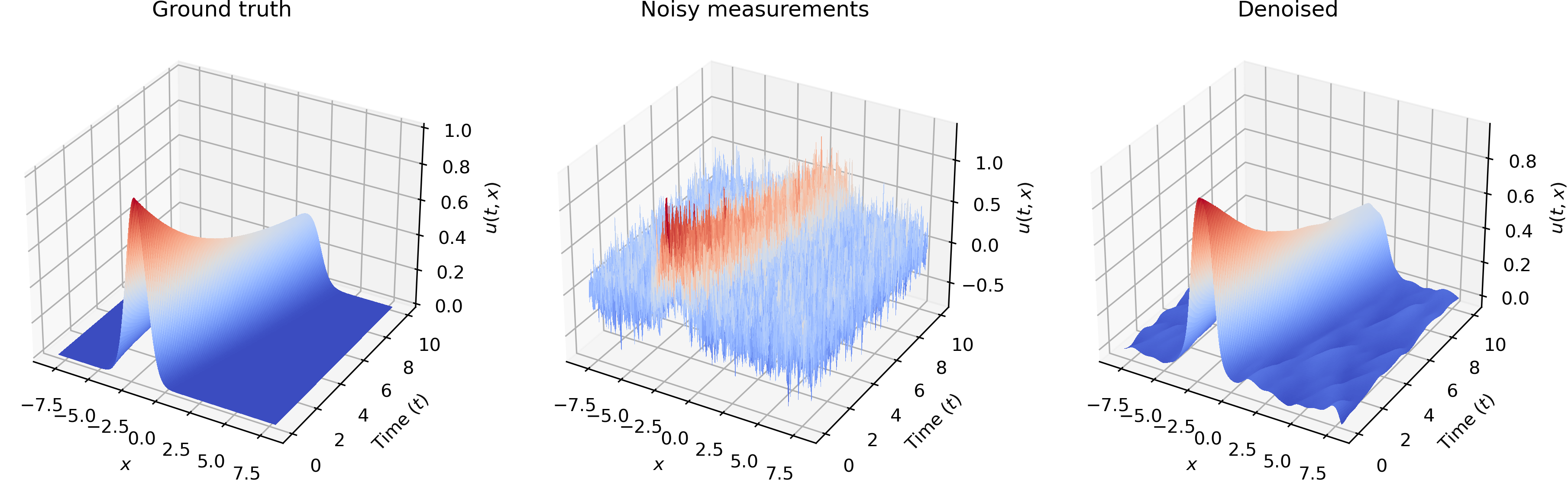}
	\includegraphics[width = 0.2\textwidth, trim = 10cm 0cm 0cm 0cm, clip]{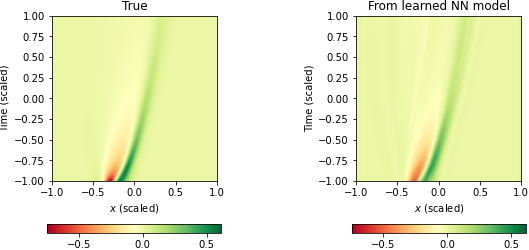}
	\rotatebox{90}{\hspace{1cm}$20\%$ noise}
	\caption{Burgers' equation: The figures show performance of the proposed framework to recover the original spatio-temporal data for various levels of noise. The first and second columns are the noisy and de-noised measurements, and the last column is the prediction of the vector field using the learning convolutional neural network. }
	\label{fig:burgers_noise}
\end{figure}

Since the data has spatial correlations, we make use of convolutional neural networks to learn the vector field of $\bu$, instead of a classical MLP as shown in \Cref{fig:NN_archi}. Thus, we build an MLP for the implicit representation and a CNN with details given in \Cref{tab:NN_info_PDE}. Once we train the network, in \Cref{fig:burgers_noise}, we plot the performance of the proposed approach to de-noise the spatial-temporal data for an increasing level of noise. We observe that the proposed methodology is able to recover the data faithfully even with significant noise in data. Furthermore, in the last columns of  \Cref{fig:burgers_noise}, we observe the approximating capability of the convolutional NN for the vector field, e.g., $\tfrac{\partial }{\partial t} \bu(t,\zeta)$. We observe that the model also predicts the vector field with good accuracy.  We mark that the vector field of the clean data is estimated using a finite-difference scheme on the clean data since the true function is not known to us. 

\subsection{Kuramoto--Sivashinsky equation}
In our last test case, we take the data of a chaotic dynamics which are obtained by simulating the Kuramoto--Sivashinsky equation which is of form:
\begin{equation}
	\dfrac{\partial}{\partial t}\bu(t,\zeta) = \bu(t,\zeta)\bu_{\zeta}(t,\zeta) + \bu_{\zeta\zeta}(t,\zeta) + \bu_{\zeta\zeta\zeta\zeta}(t,\zeta),
\end{equation}
where $\bu_{\zeta}$, $\bu_{\zeta\zeta}$, and $\bu_{\zeta\zeta\zeta\zeta}$ denote the first, second, and fourth derivatives with respect to $\zeta$. The equation explains several physical phenomena such as instabilities of dissipative  trapped ion modes in plasma, or fluctuation in fluid films, see, e.g., \cite{kuramoto1978diffusion}. We again use the data provided in \cite{rudy2017data} for the equations which is simulated using a spectral method with $1~024$ spatial grid points and $251$ time-steps. Since the dynamics present in the data is very rich, complex, and exhibits chaotic behavior, we require networks that are more expressive as compared to the previous example; the details about the networks are provided in \Cref{tab:NN_info_PDE}. 

\begin{figure}[!tb]
	\includegraphics[width = 0.25\textwidth, trim = 0cm 0cm 40cm 0cm, clip]{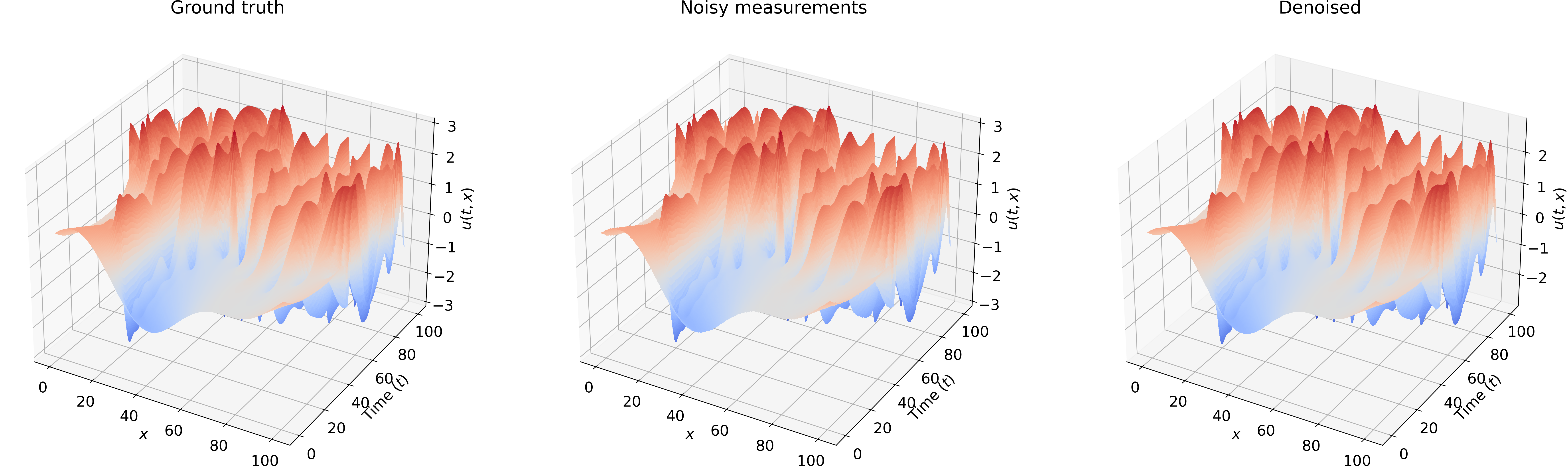}\hspace{4.4cm}
	\includegraphics[width = 0.2\textwidth, trim = 0cm 0cm 10cm 0cm, clip]{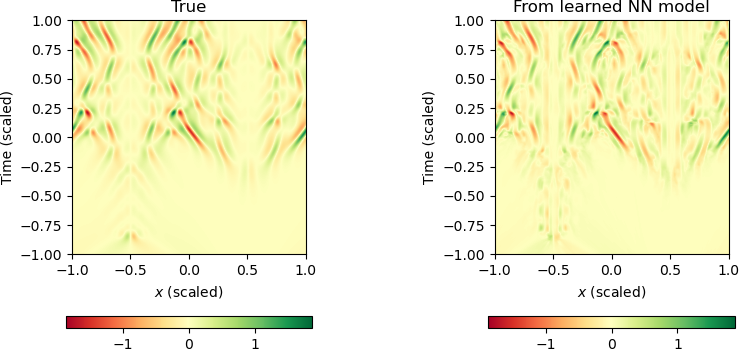} \rotatebox{90}{\hspace{1cm}clean data}
	
	\includegraphics[width = 0.5\textwidth, trim = 20cm 0cm 0cm 0cm, clip]{./Figures/KS/measurements_sampling1_noiselevel0.01_denoise.png}
	\includegraphics[width = 0.2\textwidth, trim = 10cm 0cm 0cm 0cm, clip]{./Figures/KS/measurements_sampling1_noiselevel0.01_grad}
	\rotatebox{90}{\hspace{1cm}$1\%$ noise}
	
	\includegraphics[width = 0.5\textwidth, trim = 20cm 0cm 0cm 0cm, clip]{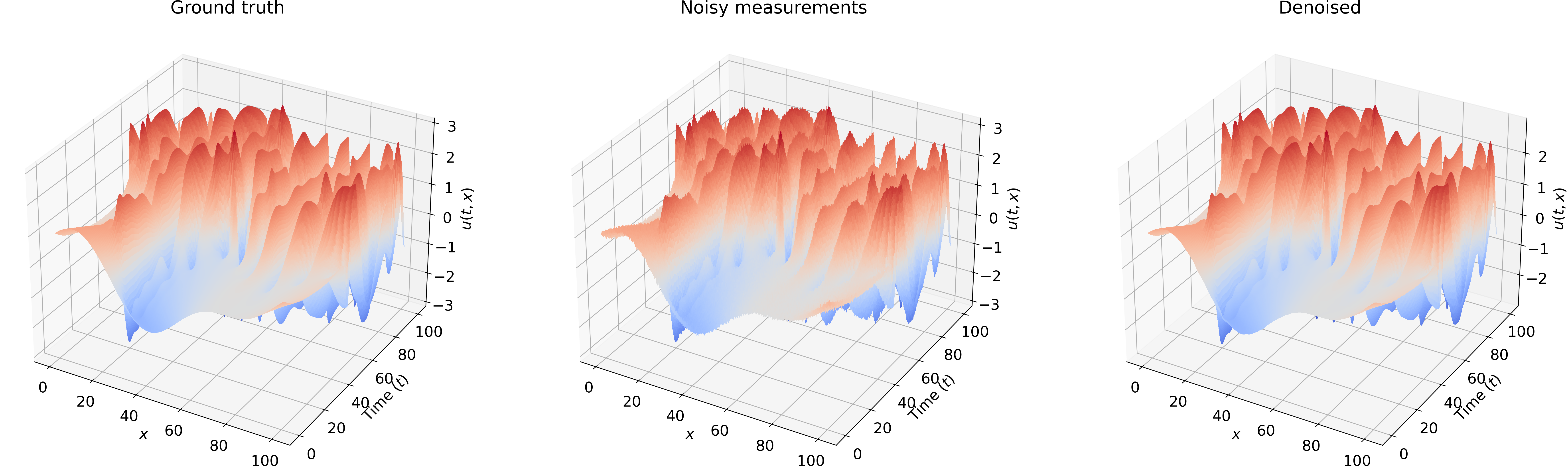}
	\includegraphics[width = 0.2\textwidth, trim = 10cm 0cm 0cm 0cm, clip]{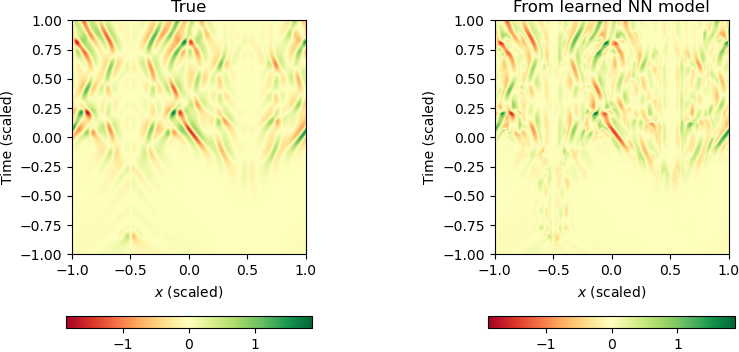}
	\rotatebox{90}{\hspace{1cm}$5\%$ noise}
	
	\includegraphics[width = 0.5\textwidth, trim = 20cm 0cm 0cm 0cm, clip]{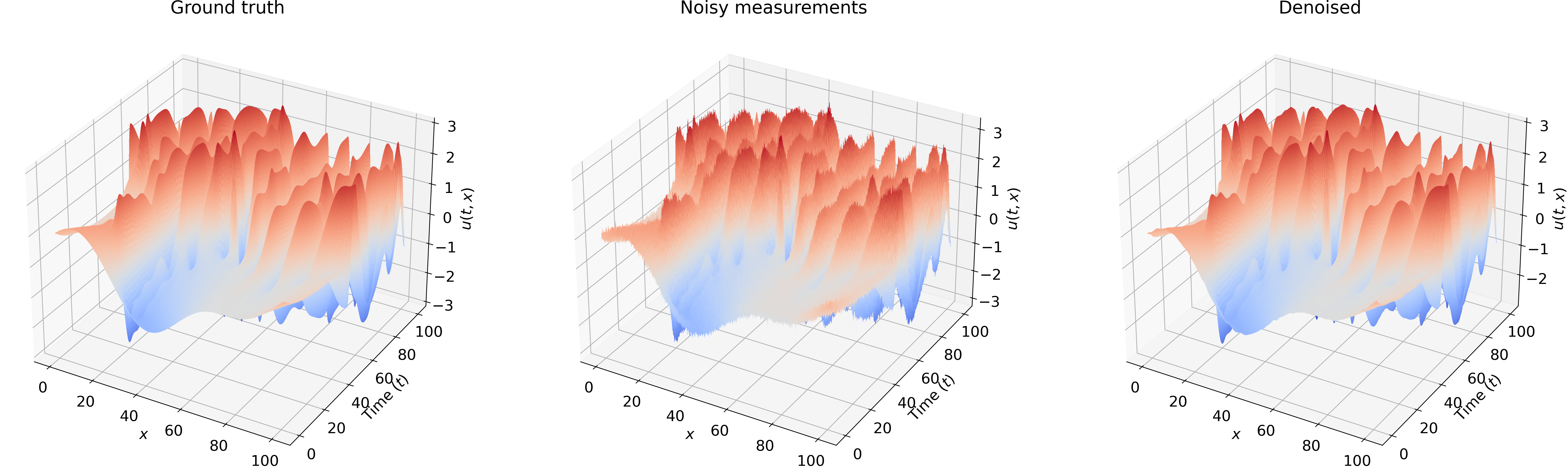}
	\includegraphics[width = 0.2\textwidth, trim = 10cm 0cm 0cm 0cm, clip]{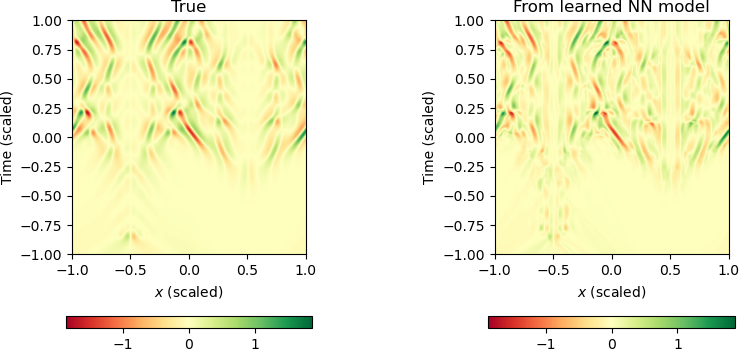}
	\rotatebox{90}{\hspace{1cm}$10\%$ noise}
	
	\includegraphics[width = 0.5\textwidth, trim = 20cm 0cm 0cm 0cm, clip]{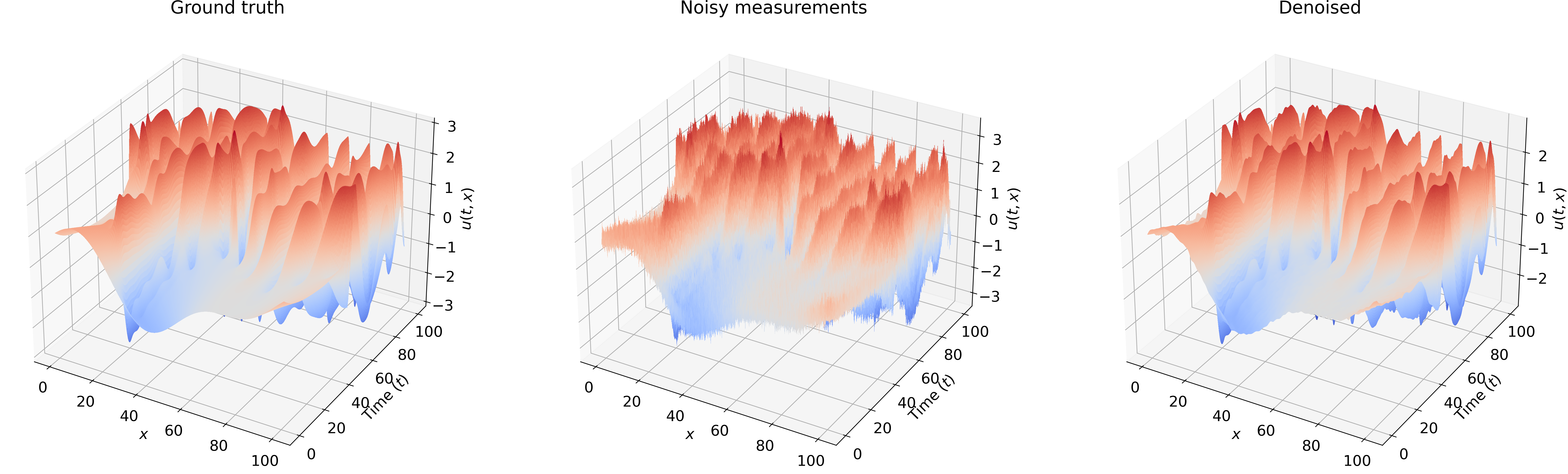}
	\includegraphics[width = 0.2\textwidth, trim = 10cm 0cm 0cm 0cm, clip]{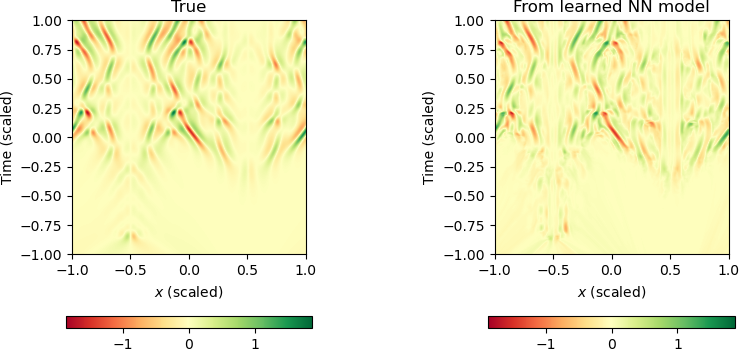}
	\rotatebox{90}{\hspace{1cm}$20\%$ noise}
	\caption{Kuramoto--Sivashinsky equation: The figure shows the noisy measurements (the left column), de-noised measurements (middle column), and the vector field approximation in the  right column.}
	\label{fig:ks_noise}
\end{figure}

In \Cref{fig:ks_noise}, we report the ability of our method to remove the noise from the spatial-temporal data.  We observe that the proposed methodology profoundly removes noise from the data. Also, the vector field is approximated very well using the learned CNN (see the last column of the figure). The vector field of the clean data is computed using a finite difference method.  We draw particular attention to the last row of \Cref{fig:ks_noise}. The algorithm recovers several minor details that are damaged due to the presence of a high-level noise ($20\%$). 

\section{Discussion}
In this work, we have presented a new paradigm for learning dynamical models from highly noisy  (spatial)-temporal measurement data. Our framework blends powerful approximation capabilities of deep neural networks with a numerical integration scheme, namely the fourth-order Runge-Kutta scheme. 
The proposed scheme involves two networks to learn an implicit representation of the measurement data and of the vector field. 
These networks are combined by enforcing that the output of the implicit network respects the integration scheme.
Furthermore, we highlight that the proposed approach can readily handle arbitrary sampled points in space and time. In fact, the dependent variables need not be collected at the same time and the exact location.  This is because we first construct an implicit representation of the data that do not require data to be of a particular structure. 

We note that the approach becomes computationally expensive when the spatial dimension increases. Indeed, it becomes impracticable when the data are collected for 2D or 3D space. A large system parameter space imposes additional challenges. However, we know that the dynamics often lie in a low-dimensional manifold. Therefore, in our future work, we aim to utilize the concept of low-dimensional embedding to make learning computationally more efficient. 
Furthermore, we learn a dynamic model as a black-box neural network. Hence, interpretability and generalizability remain opaque. In the future, it could be interesting to combine or use the de-noised data with sparse or symbolic regression, as, e.g., in \cite{rudy2017data,cranmer2020discovering,both2021deepmod} to obtain an analytic expression for a (partial) differential equations explaining the data.

\addcontentsline{toc}{section}{References}
\bibliographystyle{IEEEtran}
\bibliography{mybib}

\end{document}